\documentclass[acmtog, authorversion, nonacm]{acmart}

\usepackage{algorithm}
\usepackage{algorithmic}
\usepackage{calc}
\usepackage{graphicx}
\usepackage{subfigure}
\usepackage{multirow}
\usepackage{listings}
\usepackage{xcolor}
\usepackage{tcolorbox}

\definecolor{acmpurple}{HTML}{6D1D7D}

\definecolor{cancel_code}{HTML}{FFCCCC}
\definecolor{add_code}{HTML}{EBF1DD}
\newtcbox{\cancelbox}{on line, colframe=white, colback=cancel_code, boxrule=0pt, boxsep=0pt, top=0pt, bottom=0pt, right=4pt, left=4pt}
\newtcbox{\addbox}{on line, colframe=white, colback=add_code, boxrule=0pt, boxsep=0pt, top=0pt, bottom=0pt, right=4pt, left=4pt}
\newtcbox{\cancelboxnospace}{on line, colframe=white, colback=cancel_code, boxrule=0pt, boxsep=0pt, top=0pt, bottom=0pt, right=0pt, left=0pt}
\newtcbox{\addboxnospace}{on line, colframe=white, colback=add_code, boxrule=0pt, boxsep=0pt, top=0pt, bottom=0pt, right=0pt, left=0pt}
\newcommand{\cancelled}[1]{\cancelbox{\makebox[\widthof{+}]{-} #1}}
\newcommand{\added}[1]{\addbox{+ #1}}

\AtBeginDocument{
  }

\begin{document}

\title{Fast and Accurate Neural Rendering Using Semi-Gradients}

\author{In-Young Cho}
\email{ciy405x@krafton.com}
\author{Jaewoong Cho}
\email{jwcho@krafton.com}
\affiliation{
    \institution{KRAFTON}
    \city{Seoul}
    \country{Republic of Korea}
    \postcode{06142}
}

\renewcommand{\shortauthors}{Cho and Cho}

\begin{abstract}
We propose a simple yet effective neural network-based framework for global illumination rendering. Recently, rendering techniques that learn neural radiance caches by minimizing the difference (i.e., residual) between the left and right sides of the rendering equation have been suggested. Due to their ease of implementation and the advantage of excluding path integral calculations, these techniques have been applied to various fields, such as free-viewpoint rendering, differentiable rendering, and real-time rendering. However, issues of slow training and occasionally darkened renders have been noted.
We identify the cause of these issues as the bias and high variance present in the gradient estimates of the existing residual-based objective function. To address this, we introduce a new objective function that maintains the same global optimum as before but allows for unbiased and low-variance gradient estimates, enabling faster and more accurate training of neural networks. In conclusion, this method is simply implemented by ignoring the partial derivatives of the right-hand side, and theoretical and experimental analyses demonstrate the effectiveness of the proposed loss.
\end{abstract}



\begin{teaserfigure}
\centering
\includegraphics[width=0.734\textwidth]{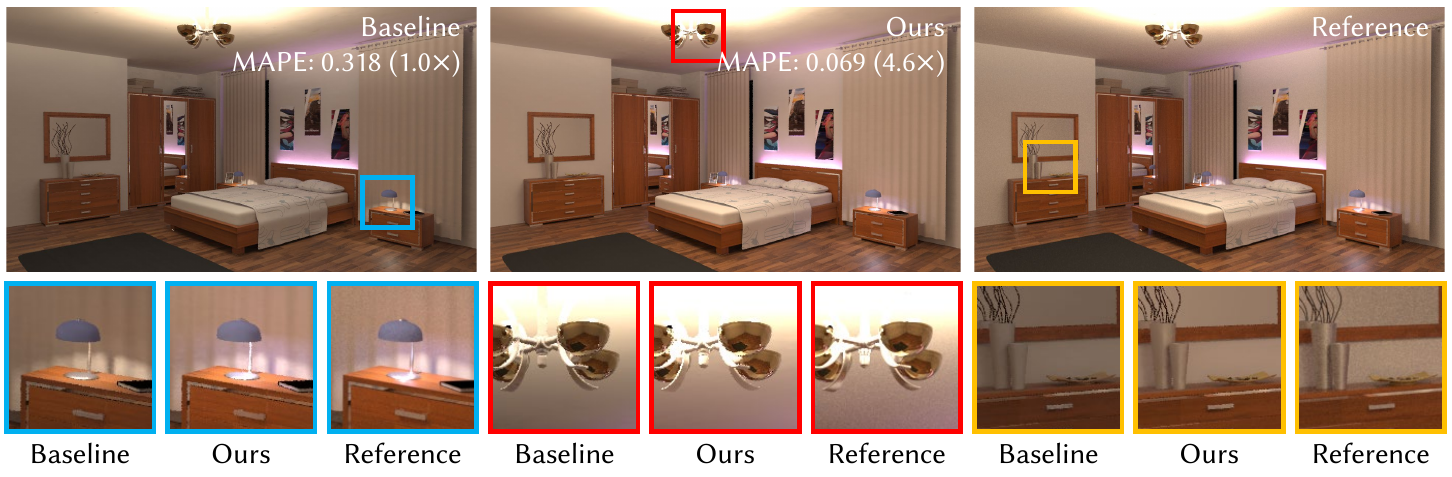}
\includegraphics[width=0.261\textwidth]{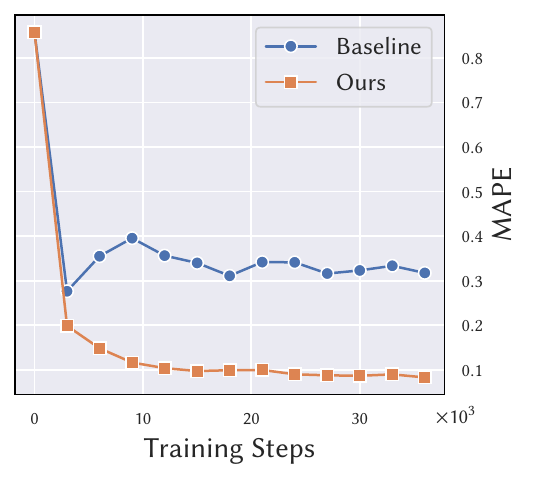}
\caption{
Rendering residual-based optimization results of the baseline, based on Neural Radiosity~\cite{hadadan2021neural}, and the proposed method for neural rendering. Previous approaches minimize the L2 distance between the left-hand side (i.e., outgoing radiance) and the right-hand side (i.e., the sum of emitted and reflected radiance) of the rendering equation.
We discover that the bias and high variance of the baseline gradient estimator lead to poor convergence, and we propose a partial derivative-based optimization method to resolve them.
The graph on the right shows the change in image-space distance from the reference as training progresses.
}
\label{fig:teaser}
\end{teaserfigure}

\maketitle

\section{Introduction}
\label{sec:introduction}

Neural network-based methods have recently received remarkable attention in global illumination rendering domain. This is primarily due to the scalability and high fidelity of neural networks, which help alleviate the computational burden and noise associated with traditional Monte Carlo (MC) methods. Among these methods, some apply neural networks to the precomputation or caching processes that have long been common in rendering.

In particular, \citet{hadadan2021neural, muller2021real} replace the outgoing radiance field on both the left-hand side (LHS) and right-hand side (RHS)---the sum of the emitted and reflected radiance---of the rendering equation with a neural network. They introduce objective functions that minimize the L2 distance between the two. These approaches use a neural network to approximate the RHS, employing fewer ray tracing iterations instead of extensive path tracing. This not only makes the approach appealing for reducing memory incoherent operations during rendering but also during the learning phase of neural caches.

\begin{figure}[]
\centering
\includegraphics[width=\linewidth]{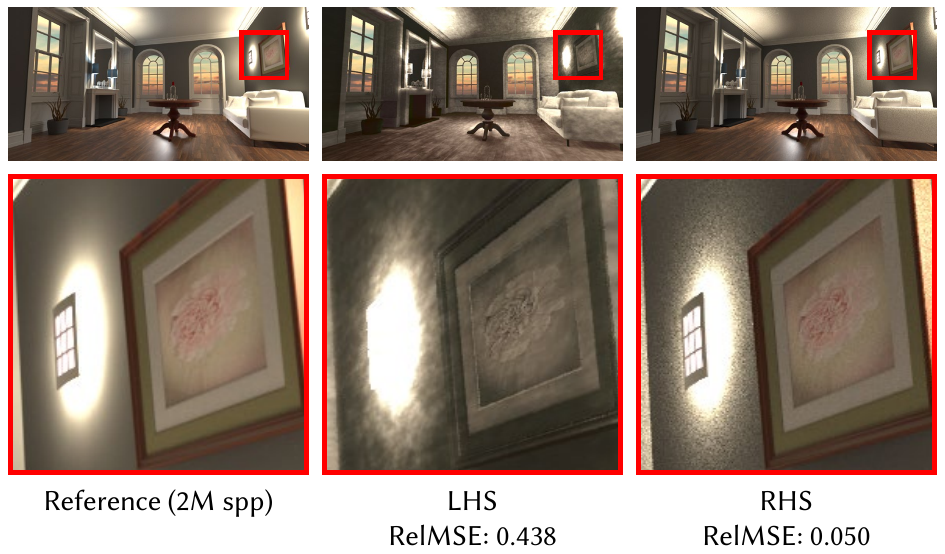}
\caption{
Example to motivate our partial derivative-based optimization for faster convergence. From the start of training (100 iterations, 14 seconds) of the baseline (Sec.~\ref{preliminaries}), the RHS quickly converges to the reference, while the LHS lags behind with issues such as faint colors and brightness overshoot on lampshades. As such, we ignore the partial derivative with respect to the RHS, focusing on learning the LHS. For visualization, the LHS and the RHS are shown at 4 and 1024 samples-per-pixel (spp), respectively.
}
\label{fig:motivating_example}
\end{figure}

While there have been various applications of this technique to real-time rendering or differentiable rendering~\cite{hadadan2023inverse, coomans2024real}, substantial dialogue regarding methods to accelerate the training of neural networks remains scarce. Meanwhile, we observe that during training using rendering residuals, the RHS converges much faster than the LHS (Fig.~\ref{fig:motivating_example}). Consequently, we hypothesize that the existing loss, which optimizes the LHS and RHS to get closer to each other, might hinder convergence. This happens because the RHS, which is already close to the reference, is updated towards the LHS, which is still far from the reference.

Therefore, we theoretically derive that using only the partial derivatives for the LHS can still lead neural caches to converge to the solution of the rendering equation, and we define this as the \textit{semi-gradient} method. Our results, also exemplified in Fig.~\ref{fig:teaser}, show that in comparisons of identical training iterations with the baseline~\cite{hadadan2021neural}, the error to the reference decreases by an average of 8.8 times across various scenes, and training time is reduced by 25-30\%.

Through theoretical and experimental analysis, we demonstrate that the stochasticity of MC integration (e.g., RHS) and its noisy derivatives cause gradient bias and high variance. This issue has also been pointed out in the differentiable rendering domain~\cite{azinovic2019inverse, nimier2020radiative}. In contrast, our proposed method effectively mitigates this error in forward rendering by removing the partial derivatives for the RHS, the main source of the error.
\section{Related Work}

\subsection{Neural Caches for Global Illumination}

Recently, deep learning-based caching techniques for rendering have been applied in various ways. Some methods employ supervised learning techniques, training neural networks using pre-collected reference renders and then utilizing the networks to render other scenes. \citet{zhu2021neural,rainer2022neural} use neural caches to learn objects illuminated by environment maps and complex luminaires, respectively. \citet{huang2024online, dong2023neural} design neural networks that decode parametric mixtures for path guiding.

Meanwhile, \textit{self-training} methods, which train neural caches to restore radiance fields without using ground-truth renders, are also actively discussed. To achieve this, \citet{hadadan2021neural,muller2021real} use the rendering equation as a strong constraint that the radiance field must satisfy, in free-viewpoint rendering and real-time rendering, respectively. Since the outgoing radiance field (at equilibrium) always represents the sum of emitted and reflected radiance, the neural cache is trained to minimize the residual of the rendering equation. This discussion has been extended to real-time dynamic scene rendering and inverse rendering by \citet{hadadan2023inverse,coomans2024real} respectively.

We focus on improving the convergence of this intriguing technique that allows learning caches without target radiance. Specifically, \citet{hadadan2021neural,hadadan2023inverse} use the gradient for the RHS in optimization, while \citet{muller2021real,coomans2024real} deactivate it. However, clear discussions supporting these choices have been scarce. Additionally, the reference implementation of \citet{hadadan2021neural} acknowledges the idea of deactivation, but the paper does not study this choice. Therefore, we revisit the losses and support the semi-gradient method through experimental and theoretical analysis (Sec.~\ref{sec:discussion}).

\subsection{Gradient Estimation for Learning and Rendering}

Gradient estimation is crucial for optimization techniques that rely on stochastic gradient descent (SGD), such as neural caching, differentiable rendering, and deep learning.

Differentiable rendering also aims to estimate accurate image gradients to optimize scene parameters. \citet{azinovic2019inverse, nimier2020radiative} highlight that gradient estimates can be biased when optimizing with MC samples, potentially distorting the converged scene parameters. To address this, methods such as using two uncorrelated samples for the rendering integral and its derivatives~\cite{azinovic2019inverse, nimier2020radiative, vicini2021path}, employing dual-buffer techniques for the L2 loss~\cite{deng2022reconstructing, pidhorskyi2022depth}, and exploiting variance reduction techniques~\cite{zhang2021antithetic, nicolet2023recursive, wang2023amortizing, fischer2023plateau, balint2023joint} have been proposed.

Reinforcement learning (RL) learns optimal actions by minimizing the residual of the Bellman integral equation~\cite{baird1995residual}. However, in real-world scenarios, using uncorrelated samples is not always possible. Thus, empirical methods that utilize only the semi-gradient of the Bellman equation are employed~\cite{mnih2015human}. Despite the wide adoption of the semi-gradient method, a clear theoretical analysis of its convergence and effectiveness remains open~\cite{zhang2019deep, yin2022experimental, sharifnassab2023toward}.

We show that gradient estimates for the rendering residual are also biased due to MC samples. Intriguingly, while the dual-buffer method effectively addresses this bias, the convergence of neural caches still remains slow. We found that this is due to the high variance of the RHS partial derivatives, which requires spending 16 times more time to extract 8 times more samples to resolve the issue (Sec.~\ref{sec:discussion}). However, we demonstrate that completely removing the RHS partial derivatives (\textit{semi-gradient}) is more effective for convergence.
\section{Neural Rendering Using Semi-Gradients}

\subsection{Preliminaries and Baseline Method}
\label{preliminaries}
The rendering equation~\cite{kajiya1986rendering} serves as the cornerstone of physically-based rendering, representing the spatio-directional radiance field at equilibrium \(L^*\):
\begin{equation} \label{eq:1}
\begin{split}
L^*(x, \omega) & = E(x, \omega) \\
& + \int_{\Omega} L^*(r(x, \omega_{\text{in}}), -\omega_{\text{in}}) f_s(x, \omega, \omega_{\text{in}}) |n^\text{T} \omega_{\text{in}}| d\omega_{\text{in}}.
\end{split}
\end{equation}
In short, the equation states that the radiance \(L^*\) outgoing from a point \(x\) in a direction \(\omega\) is always due to two sources: 1) emission \(E(x, \omega)\) at the surface (e.g., in the case of a light source), and 2) reception from the surroundings. \(f_s(x, \omega, \omega_{\text{in}})\) is the bidirectional scattering distribution function (BSDF), \(|n^\text{T} \omega_{\text{in}}|d\omega_{\text{in}}\) is the differential of the projected solid angle between a surface normal \(n\) and an incident direction \(\omega_{\text{in}}\), and \(r\) is the ray tracing operation.

To train neural radiance caches that satisfy the rendering equation, \citet{hadadan2021neural, hadadan2023inverse, coomans2024real, mueller2022instant} sample \((x, \omega)\) on meshes and a unit hemisphere in each training iteration. They approximate the LHS and RHS using neural networks and train the network by minimizing the L2 norm of the residuals. The loss is formulated as follows:
\begin{equation}
\mathcal{L}(\theta) = \frac{\|L_\theta - R_\theta\|^2}{\|\text{sg}(L_\theta)\|^2 + \epsilon}.
\label{eq:nerad_loss}
\end{equation}
Here, \(L_\theta\) is a neural network parameterized by \(\theta\) that approximates \(L^*\), and \(R_\theta\) represents the RHS of the rendering equation with \(L_\theta\) substituted for \(L^*\) in the integrand. For brevity, the \((x, \omega)\) notation is omitted. The \(\text{sg}(\cdot)\) denotes the stop-gradient operation, which ignores the gradient during the backward pass of automatic differentiation:
\begin{equation}
\begin{split}
\text{sg}(x) &= x, \\
\nabla_x \text{sg}(x) &= 0.
\end{split}
\end{equation}
Note that this is a notation for implementation convenience, not a mathematically rigorous operation.

To approximate the above loss over the entire mesh and hemisphere space, \citet{hadadan2021neural} uniformly sample \((x, \omega)\). For each \((x, \omega)\), the MC estimate for \(R_\theta\) is calculated as follows to compute \(\mathcal{L}(x, \omega; \theta)\):
\begin{equation}
\langle R_\theta \rangle = E + \frac{1}{M}\sum_{j=1}^{M}\frac{L_\theta(r(\omega_{\text{in},j}), -\omega_{\text{in},j}) f_s(\omega_{\text{in},j}) |n^\text{T} \omega_{\text{in},j}|}{p(\omega_{\text{in},j})},
\label{eq:<R>}
\end{equation}
where \(\langle \cdot \rangle\) denotes an estimator (whether it is biased or not), \(M\) is the incident sample count and \(p(\omega_{\text{in}}|x, \omega)\) is the probability density function (PDF) from which an incident direction is sampled. As long as the support of the PDF includes the support of the numerator, this MC estimate is an unbiased estimate of \(R_\theta\).

Finally, the original loss is approximated through the aforementioned sampling as follows:
\begin{equation}
\langle \mathcal{L}(\theta) \rangle_{\sf NR} = \frac{\|L_\theta - \langle R_\theta \rangle\|^2}{\|\text{sg}(L_\theta)\|^2 + \epsilon}.
\label{eq:<L>}
\end{equation}
Here, {\sf NR} stands for Neural Radiosity, the pioneering work by \citet{hadadan2021neural}. The gradient estimator for optimizing this loss is approximated as follows:
\begin{equation}
    \begin{split}
    \nabla_\theta \langle \mathcal{L(\theta)}\rangle_{\sf NR} &=
    \nabla_\theta L_\theta \cdot \partial_{L_\theta} \langle \mathcal{L}(\theta)\rangle_{\sf NR} 
    + \nabla_\theta \langle R_\theta \rangle \cdot \partial_{\langle R_\theta \rangle} \langle \mathcal{L}(\theta)\rangle_{\sf NR}  \\
    &= \nabla_\theta L_\theta \cdot \frac{2(L_\theta - \langle R_\theta \rangle)}{\|L_\theta\|^2 + \epsilon} 
    + \nabla_\theta \langle R_\theta \rangle \cdot \frac{-2(L_\theta - \langle R_\theta \rangle)}{\|L_\theta\|^2 + \epsilon}.
    \end{split}
    \label{eq:biased_gradient_estimator}
\end{equation}
We note that \(\text{sg}(\cdot)\) disappears in gradient evaluation. In Sec.~\ref{sec:discussion}, we show that this gradient estimator is a \textit{biased} estimator of the gradient of Eq.~(\ref{eq:nerad_loss}). We reveal that it, along with high variance, hinders convergence. The common training procedure is summarized in Algorithm~\ref{algo:faster_neural_radiosity} with the \cancelboxnospace{red background}. 

\subsection{Semi-Gradient Method}

\begin{algorithm}[t!]
\caption{Color-coding the changes of our method compared to the baseline}
\begin{algorithmic}[1]
\REQUIRE Initialize network parameters \(\theta\) and the learning rate \(\eta\)
\WHILE{not converged}
    \STATE Sample location vectors \( \{x_j\}_{j=1}^{N} \) uniformly on meshes
    \STATE Sample the outgoing directions \( \{\omega_{j}\}_{j=1}^{N} \) uniformly over 
        \STATE \quad a hemisphere
    \FOR{each \( (x_j, \omega_{j}) \)}
        \STATE Sample incident directions \( \{\omega_{\text{in}, j, k} | k = 1, \ldots, M\} \)
        \STATE \quad uniformly over a hemisphere
    \ENDFOR
    \STATE \cancelled{Evaluate \(  \langle\mathcal{L} (\theta)\rangle_{\sf NR} \) via the above samples}
    \STATE \added{Evaluate \(  \langle\mathcal{L} (\theta)\rangle_{\sf SG} \) via the above samples}
    \STATE \cancelled{\( \theta \leftarrow \theta - \eta \cdot \nabla_\theta \left\langle\mathcal{L} (\theta)\right\rangle_{\sf NR}\)}
    \STATE \added{\( \theta \leftarrow \theta - \eta \cdot \nabla_\theta \left\langle\mathcal{L} (\theta)\right\rangle_{\sf SG}\)}
\ENDWHILE
\RETURN \( \theta \)
\end{algorithmic}
\label{algo:faster_neural_radiosity}
\end{algorithm}

We previously raised the question in Sec.~\ref{sec:introduction} and Fig.~\ref{fig:motivating_example} whether it is necessary to update \(R_\theta\) towards \(L_\theta\). Indeed, the RHS partial derivative in Eq.~(\ref{eq:biased_gradient_estimator}) can be seen as instructing the network to pull the RHS estimates towards the LHS. To prevent this, we replace the loss in the baseline training procedure with the following:
\begin{equation}
\langle\mathcal{L}(\theta)\rangle_{\sf SG} = \frac{\|L_\theta - \text{sg}(\langle R_\theta \rangle)\|^2}{\|\text{sg}(L_\theta)\|^2 + \epsilon}.
\label{eq:semi_gradient_loss}
\end{equation}
Thanks to the stop-gradient operation introduced in the numerator, \(\partial_{\langle R_\theta \rangle} \langle\mathcal{L}(\theta)\rangle_{\sf SG} = 0\), achieving our intended goal. Here, {\sf SG} stands for the \textit{semi-gradient} method, as opposed to the full-gradient method. Therefore, the gradient estimator for optimizing this loss is evaluated as follows:
\begin{equation}
\nabla_\theta \langle \mathcal{L}(\theta) \rangle_{\sf SG} = \nabla_\theta L_\theta \cdot \frac{2(L_\theta - \langle R_\theta \rangle)}{\|L_\theta\|^2 + \epsilon}. 
\label{eq:gradient_of_semi_gradient}
\end{equation}

However, for radiance caching, it must be ensured that the neural cache updated along this new gradient converges to the solution of the rendering equation. In Appendix~\ref{appendix:convergence}, we theoretically derive that convergence can be guaranteed for scenes satisfying the energy absorption condition~\cite{neumann1995radiosity}. It is important to note that due to the mathematical shortcomings in the definition of the stop-gradient operation, the convergence of this L2-form loss cannot be prematurely guaranteed. For formalism regarding the expression and evaluation of the stop-gradient operation, refer to \citet{scibior2021differentiable}.

Additionally, although semi-gradient formulations have been utilized in reinforcement learning through temporal difference methods, as well as in neural rendering~\cite{coomans2024real, mueller2022instant}, a distinct theoretical explanation of the effectiveness has not been thoroughly explored. We provide both theoretical and experimental analyses from a light transportation perspective.

Finally, Sec.~\ref{sec:results} empirically shows that this subtle yet crucial alteration significantly enhances performance. Sec.~\ref{sec:discussion} provides theoretical and experimental explanations for the improvement. Algorithm~\ref{algo:faster_neural_radiosity} summarizes the overall training procedure with the \addboxnospace{green background}.

\subsection{Implementation}
We implement the baseline algorithm and our models following the PyTorch-based implementation by \citet{cho2023neural}. All radiance prediction networks consist of seven linear layers of 512 units with six ReLU activations; the last output of the linear layer does not go through an activation layer.

Network parameters are initialized using the Xavier uniform distribution~\cite{glorot2010understanding}. As previously noted, the training employs the Adam optimizer~\cite{kingma2014adam} with a learning rate set to \(5 \times 10^{-4}\). The batch size for \((x, \omega)\) (i.e., \(N\) in the Algorithm~\ref{algo:faster_neural_radiosity}) is set to \(2^{14}\) at the beginning, and for each \((x, \omega)\), 32 incident directions are sampled (i.e., \(M\)). A radiance field network is trained for 36,000 steps separately for each scene. The learning rate is reduced by a third every 12,000 steps. We utilize the emission reparameterization trick proposed in the original work~\cite{hadadan2021neural} to ensure stable training.

Training is performed on a single NVIDIA A100 GPU, with models using 7GB or less of VRAM. Training times vary depending on scene complexity and models, ranging from 40 minutes to 3.5 hours.

We use Mitsuba~3~\cite{jakob2022mitsuba3}, which integrates smoothly with PyTorch for all rendering tasks in Algorithm~\ref{algo:faster_neural_radiosity}. As suggested by \citet{hadadan2021neural, dong2023neural}, neural networks are fed with not only position and direction vectors but also texture, normal, and multi-resolution encoding inputs. 

\paragraph{Variance reduced gradient estimator.}
To expedite initial convergence, we integrate the neural path guiding technique~\cite{dong2023neural} into our framework. Additional neural caches are employed for sampling incident directions, comprising four linear layers of 256 units with three activations. These tiny caches result in negligible overhead compared to the baseline BSDF sampling. During 1,000 to 5,000 iterations, this module is trained alongside radiance caches, while held fixed for the remaining iterations.

This path guiding model operates with next event estimation for multiple importance sampling, aiding in the reduction of the variance of the RHS estimates and subsequently decreasing the variance of the semi-gradient estimator (Eq.~(\ref{eq:gradient_of_semi_gradient})). This reduction is often instrumental in mitigating visual artifacts during early training stages and accelerating error reduction~\cite{wang2013variance, liu2020adam}. However, the variance-reduced gradient does not impact the error at final convergence, the primary focus of our discussion. Please refer to Fig.~\ref{fig:converges_faster} in the Appendix for the ablation study on variance reduction.

\subsection{Speedup Tricks Across Comparative Models}
\label{sec:speedup_tricks}

Neural Radiosity is an innovative method, but it often requires significant training time compared to state-of-the-art MC methods. 
We apply common modifications to all models to fasten training and rendering, resulting in approximately 2-fold faster processing. Specifically, we use the tiny-cuda-nn library~\cite{tiny-cuda-nn} to implement all networks. We also replace the multi-resolution feature grid with hash encoding from InstantNGP~\cite{mueller2022instant}, which is also employed by \citet{dong2023neural}. For the multi-resolution hash encoding, we set the base resolution to 2, the number of hierarchy levels to 14, the number of features per level to 2, and the hashmap size to 18 (in log scale).

Additionally, both training and rendering use mixed precision, which does not noticeably affect the quality of results. Thus, inputs like \((x, \omega)\) are rendered in 32-bit full precision and then converted to 16-bit half-precision when fed into the network. The network outputs are converted back to full precision before being ingested by the rendering engine. 

Last but not least, to fully realize the speedup of the semi-gradient technique, we emphasize that unnecessary intermediate activations should not be stored for the backward pass, for example, by using the \textit{no-grad} context manager in PyTorch~\cite{paszke2017automatic}.
\section{Results}
\label{sec:results}

\subsection{Dataset and Evaluation}

We evaluate the performance of all models mainly across seven scenes with diverse interactions between materials and lights. To assess the convergence process, rendering via radiance prediction network is performed on a view determined for each scene every 3,000 steps and compared with the reference image. Metrics such as MSE, MAPE, and RelMSE are measured at high-definition resolutions of either 960~\(\times\)~960 or 1,280~\(\times\)~720. 

The reference images were rendered using 256K samples-per-pixel (spp), and the reference videos were rendered using 4K spp for each frame, followed by post-processing with OptiX denoiser~\cite{parker2010optix} introduced in Mitsuba 3. Rendering every frame of a video with 256K spp would be too costly, so we used fewer spp and then applied denoising as post-processing for the reference videos. This approach more closely resembles the typical rendering process. All neural renderings, except RHS renderings, utilized four spp.

\begin{figure}[]
\centering
\includegraphics[width=\linewidth]{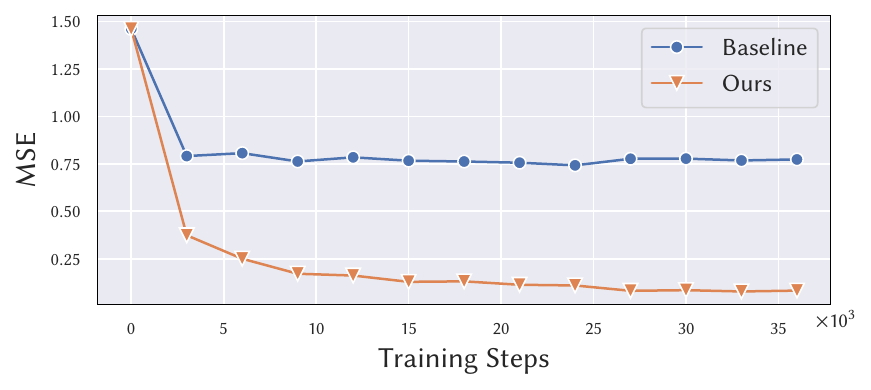}
\caption{
Equal-iteration comparisons for our method and the baseline approach. The figure displays the convergence of image error, averaged across seven scenes in Fig.~\ref{fig:superior_reconstruction}. Our method shows an average error that is 8.8 times lower.
}
\label{fig:converges_faster}
\end{figure}

\subsection{Training Efficiency Comparison}

\begin{table}[]
\centering
\caption{
Comparison of time taken per iteration during training between the baseline and our method (in seconds). Our method takes 25-30\% less time.
}
\begin{tabular}{lcc}
\hline
         & Baseline & Ours  \\ \hline
Bed      & 0.207    & 0.138 \\
Bidir    & 0.096    & 0.066 \\
Bath     & 0.208    & 0.143 \\
Door     & 0.099    & 0.071 \\
Greyroom & 0.170    & 0.128 \\
Hall     & 0.113    & 0.078 \\
TVRoom   & 0.342    & 0.240 \\ \hline
\end{tabular}
\label{tab:cost_breakdown}
\end{table}

Fig.~\ref{fig:converges_faster} demonstrates that the proposed method significantly enhances per-iteration convergence compared to the baseline. Our method achieves final errors with the reference images that are nine times lower than those of the baseline loss approach on average.

Our method reduces the per-iteration cost as well. Table~\ref{tab:cost_breakdown} shows that ours reliably reduces the time to train models by 25-30\% across all benchmarks. This efficiency is primarily due to the elimination of partial derivatives for the RHS, which obviates the need to store intermediate activations in memory.

\subsection{Rendering Fidelity Comparison}

\begin{figure}[]
\centering
\includegraphics[width=\linewidth]{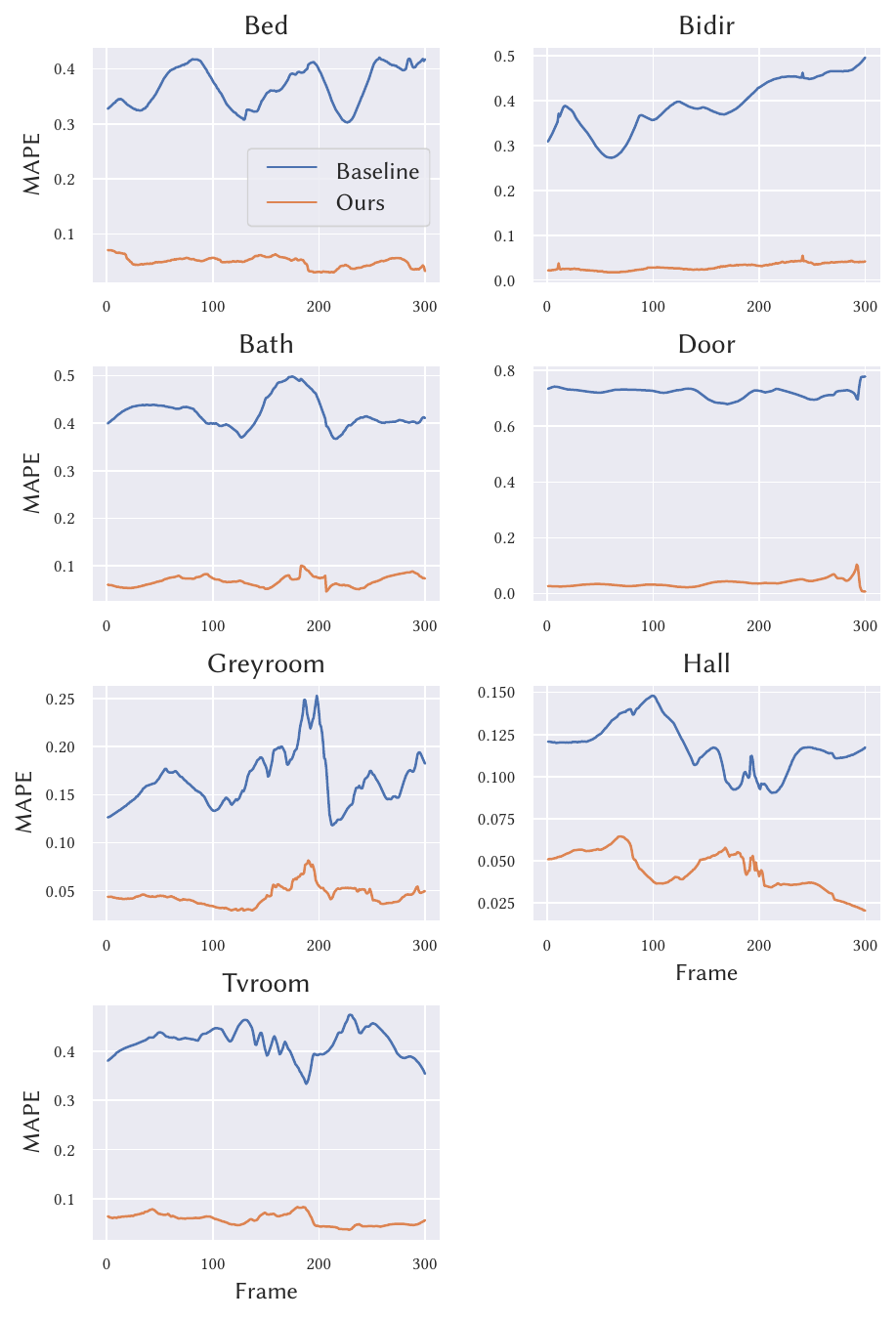}
\caption{
When rendering free-viewpoint video along a camera path in each scene after training, the error of our method is lower than the baseline in all frames. Refer to the supplementary materials for the videos.
}
\label{fig:various_scenes_and_views}
\end{figure}

Fig.~\ref{fig:superior_reconstruction} demonstrates that, once all models have fully converged after 36,000 training steps, our model renders images of higher quality both numerically and visually compared to the baseline. The baseline tends to underestimate the energy irradiated in a scene and fails to capture bright highlights, color bleeding, or other indirect illumination features. Our framework addresses these limitations. 

Lastly, Fig.~\ref{fig:various_scenes_and_views} shows that our performance is not just confined to a specific view in each scene. Consistent benefits of the proposed framework are observed in every frame of videos, rendered along camera paths traversing each scene. We observe a significant reduction in MAPE, with errors decreasing by a factor of 2 to 35 times compared to the baseline. The supplementary materials include videos for reference.
\begin{figure}[]
\centering
\includegraphics[width=\linewidth]{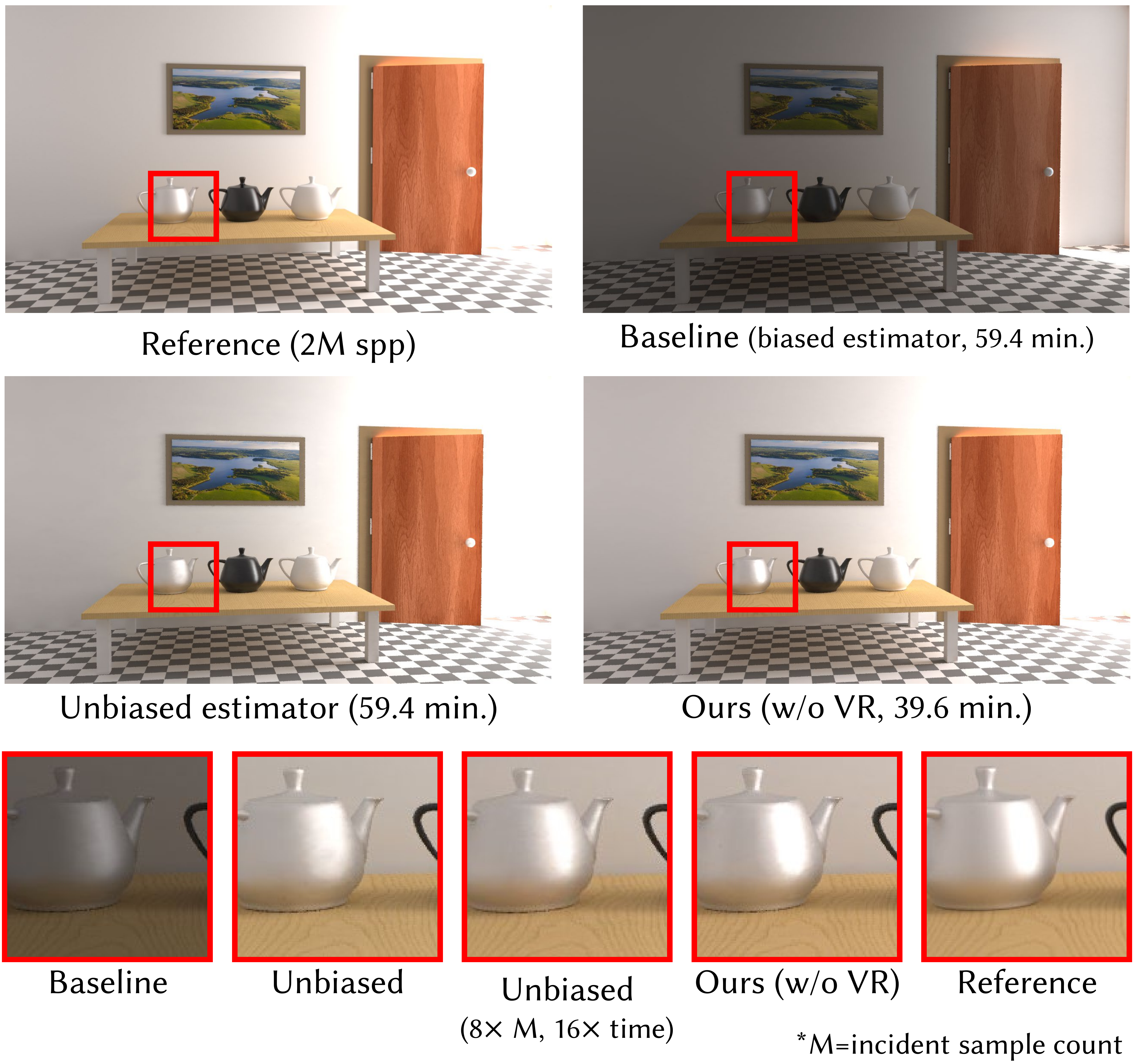}
\caption{
Visual comparison between the baseline (biased gradient estimator), dual-buffer (unbiased gradient estimator)~\cite{pidhorskyi2022depth, deng2022reconstructing}, and our methods. When using the baseline method, the resulting reconstruction is noticeably darker. The dual-buffer approach resolves the bias on gradients yet fails to reconstruct glossy reflections shapely in the same training iterations. Increasing the incident sample count eightfold (to 256), taking 16-fold time, is necessary to achieve successful restoration. Our method succeeds in reconstruction with even less time than the baseline, without the need for such excessive sampling.
{\sf VR} stands for variance reduction.
}
\label{fig:unbiased_gradient}
\end{figure}

\section{Discussion}
\label{sec:discussion}
In this section, we provide two interpretations for the faster convergence of the proposed semi-gradient method, which eliminates the partial derivative for the RHS, compared to the baseline in Eq.~(\ref{eq:biased_gradient_estimator}). Additionally, we present an ablation study.
\paragraph{\textbf{Bias}}
As demonstrated in Appendix~\ref{appendix:baseline}, the baseline is a biased estimator of the actual gradient of the loss. This bias arises from the correlation between \(\nabla_\theta \langle R_\theta \rangle\) and \(\langle R_\theta \rangle\) in the partial derivative with respect to the RHS.
Such bias ultimately stems from the stochastic nature of \(\langle R_\theta \rangle\). This bias in gradient estimation can impede the convergence to the minimum by SGD algorithm~\cite{bottou2010large}, as illustrated well by the results in Fig.~\ref{fig:unbiased_gradient}, demonstrating the erroneous convergence due to bias. 

Moreover, across various scenes depicted in Fig.~\ref{fig:superior_reconstruction}, the results consistently appear darker than the reference when using the baseline method. This occurs because as scenes darken, the variance of \(\langle R_\theta \rangle\) decreases, and so does the covariance between \(\nabla_\theta \langle R_\theta \rangle\) and \(\langle R_\theta \rangle\). Moreover, the expectation of the baseline gradient estimator is the sum of the actual gradient and the covariance between \(\nabla_\theta \langle R_\theta \rangle\) and \(\langle R_\theta \rangle\) (Eq.~(\ref{eq:biasedness_proof}) in the Appendix). Therefore, the covariance must decrease for the expectation of the gradient estimator to become sufficiently small, causing the SGD algorithm to halt.

Similar issues regarding the stochasticity of MC estimates of rendering integrals causing bias in gradients have been noted in the field of inverse rendering~\cite{azinovic2019inverse, nimier2020radiative}. We extend this discussion to forward rendering as well. 

On the contrary, while the semi-gradient loss shares the same minima as the baseline loss, its gradient estimator is not formulated as the product of \(\nabla_\theta \langle R_\theta \rangle\) and \(\langle R_\theta \rangle\), thus sidestepping the issue of bias.

\paragraph{\textbf{Variance}}
However, bias is not the sole contributing factor. When employing the dual-buffer (or half-buffer) method~\cite{deng2022reconstructing, pidhorskyi2022depth}, which provides an unbiased estimator for the gradient of the L2 loss (see Appendix~\ref{appendix:dual_buffer}), the overall brightness of the scene is well-reconstructed (see Fig.~\ref{fig:unbiased_gradient}). Nonetheless, visual aspects such as glossy reflections do not achieve the highest quality of reconstruction within 36,000 iterations. 

Therefore, we point to variance as another factor contributing to slow convergence. For both the baseline and dual-buffer methods, the RHS partial derivative involves the product of the two random variables, \(\nabla_\theta \langle R_\theta \rangle\) and \(\langle R_\theta \rangle\). Consequently, the overall variance corresponds to the product of the variances of these two variables, leading to excessive variance and hindering learning. In Fig.~\ref{fig:unbiased_gradient}, increasing the incident sample count eightfold would reduce the variance in the RHS derivative, bringing the visual quality on par with ours and the reference. However, this requires 16 times more training time. Additionally, in the simple CornellBox scene at Fig.~\ref{fig:limitations}, the variance in RHS estimates would be certainly low, so the dual-buffer method converges as well as ours. 

Meanwhile, the semi-gradient estimator, which entirely ignores the RHS derivative, is free from the burden of double variance.

\begin{figure}[]
\centering
\includegraphics[width=\linewidth]{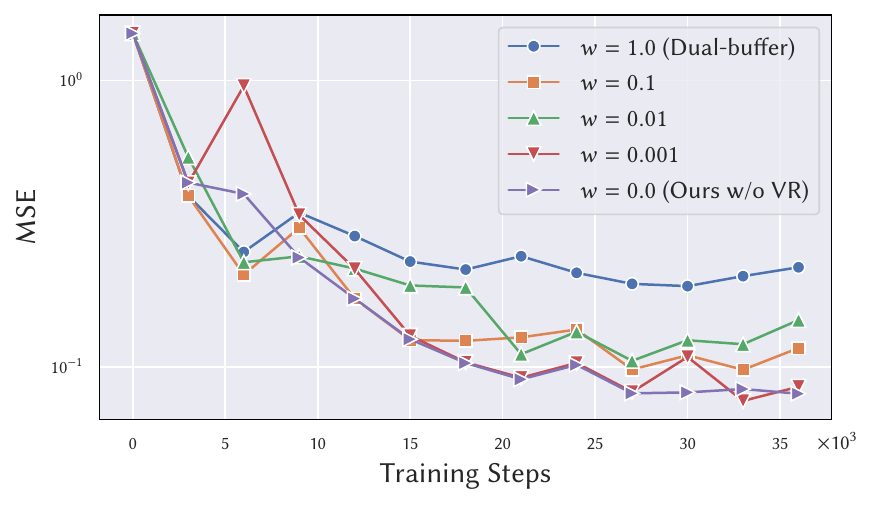}
\caption{
Linear interpolation between the gradients of the dual-buffer and our method using weight \(w\). The key difference between the two methods is whether the RHS partial derivative is included in the gradient. As the weight is reduced to 0, the influence of the RHS partial derivative diminishes, converging to the semi-gradient method, and the model's performance improves. The figure displays the convergence of image error, averaged across seven scenes in Fig.~\ref{fig:superior_reconstruction}. {\sf VR} stands for variance reduction.
}
\label{fig:detrimental_effect}
\end{figure}

\paragraph{\textbf{Ablation study}}
In addition to theoretical results, to more clearly demonstrate the advantages of the semi-gradient estimator over an unbiased gradient estimator for the baseline loss (e.g., the dual-buffer method), we train neural caches by linearly combining the two gradient estimates. When the weight \(w=1\), the gradient estimator corresponds to the gradient estimator of the dual-buffer method, and when the \(w=0\), it corresponds to the semi-gradient estimator. Please refer to Appendix~\ref{appendix:weighted} for the implementation details of the weighted loss.
In other words, by manipulating the weight associated with the RHS partial derivative, we can discern the impact on learning progression. Ultimately, Fig.~\ref{fig:detrimental_effect} demonstrates that as RHS weights decrease, performance progressively improves, thus experimentally proving the detrimental effect of RHS partial derivatives on convergence.

\section{Limitations and Future Work}
\label{sec:limitations}

\begin{figure}
\centering
\includegraphics[width=\linewidth]{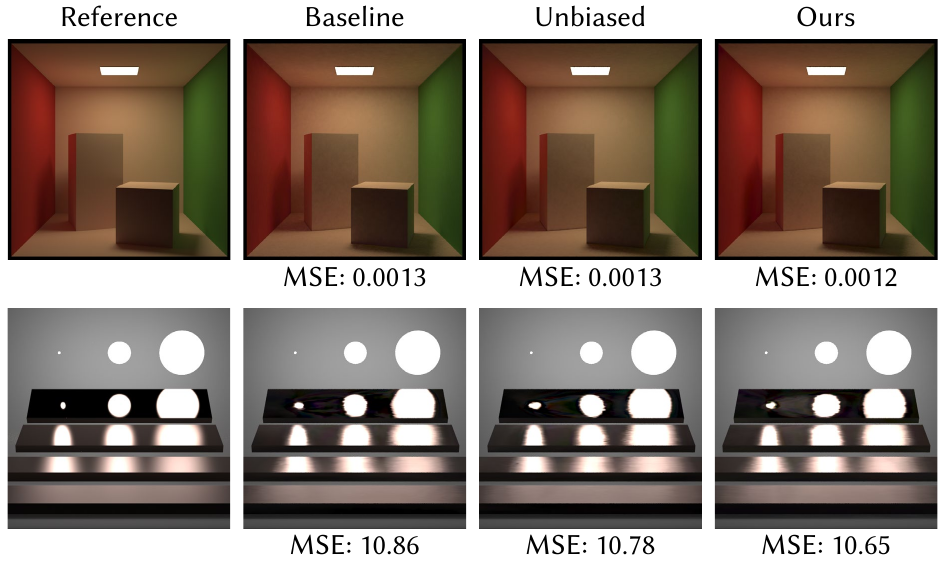}
\caption{
Examples where our approach does not provide performance advantages. 
In the first row, where the scene is simple and the stochasticity of RHS estimates is low, the issues of bias and high variance in the baseline gradient estimator are already negligible. 
In the second row, as the plates become smoother, all models struggle with reconstruction. \textit{Unbiased} refers to the model using the dual-buffer method.
}
\label{fig:limitations}
\end{figure}

Our proposed method demonstrates numerical and visual improvements in most scenes (see Fig.~\ref{fig:superior_reconstruction}), but there are certain scenes where the improvements are limited. As shown in the first row of Fig.~\ref{fig:limitations}, in simple scenes where the RHS can be accurately estimated through next event estimation, all models exhibit similar performance. Additionally, in the scenes shown in the second row, which can be easily rendered using simple path tracing, all models struggle to capture highly view-dependent reflection effects. The supplementary video of the TVRoom scene also shows similar difficulties in capturing reflections on rough glass surfaces on a TV. We aim to explore effective directional space learning techniques to overcome these challenges in future work.

Extending the semi-gradient technique to differentiable rendering is another intriguing problem. The challenges of physically-based differentiable rendering include 1) implementing and theorizing the derivative of the path integrals and 2) differentiation on noisy inputs, which hampers accurate gradient estimation. Recent studies have applied the advantages of residual training-based neural caches to differentiable rendering, excluding the need for path integral evaluation~\cite{hadadan2023inverse}. This allows derivatives of the path integral to be easily obtained using neural networks and automatic differentiation. Applying the semi-gradient technique here could avoid differentiation on noisy inputs, facilitating gradient estimation.

\begin{figure}
\centering
\begin{minipage}{\linewidth}
    \raggedright
    \includegraphics[width=0.93\linewidth]{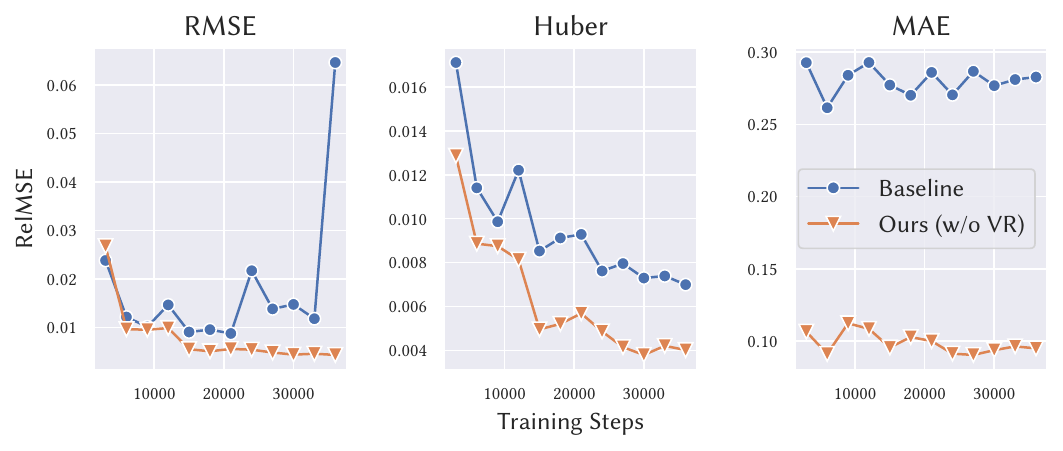}
\end{minipage}
\includegraphics[width=\linewidth]{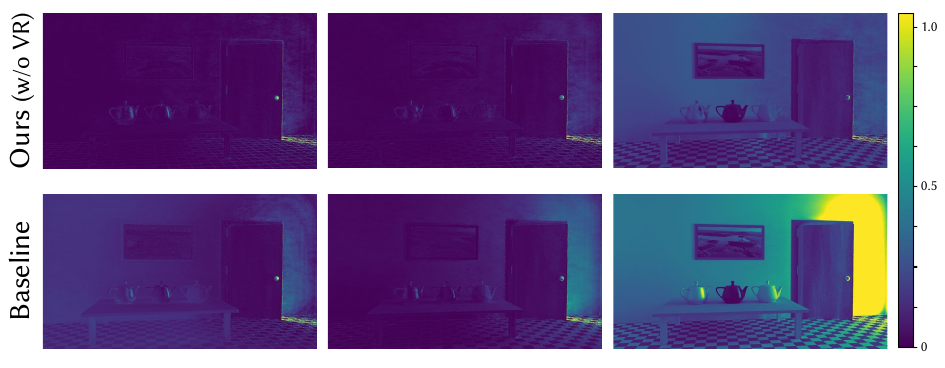}
\caption{
Visualization of the convergence and final error maps when applying the semi-gradient and baseline techniques to optimize objective functions other than the L2 loss (e.g., RMSE, Huber, MAE). The semi-gradient method consistently reduces error more effectively than the baseline, even when using non-L2 objectives, which do not necessarily guarantee unbiased convergence. For better visualization, the RelMSE maps are tone-mapped, and the 0-th step errors are omitted in the first row.
}
\label{fig:preliminary_results}
\end{figure}

\begin{figure*}[]
\centering
\includegraphics[width=\linewidth]{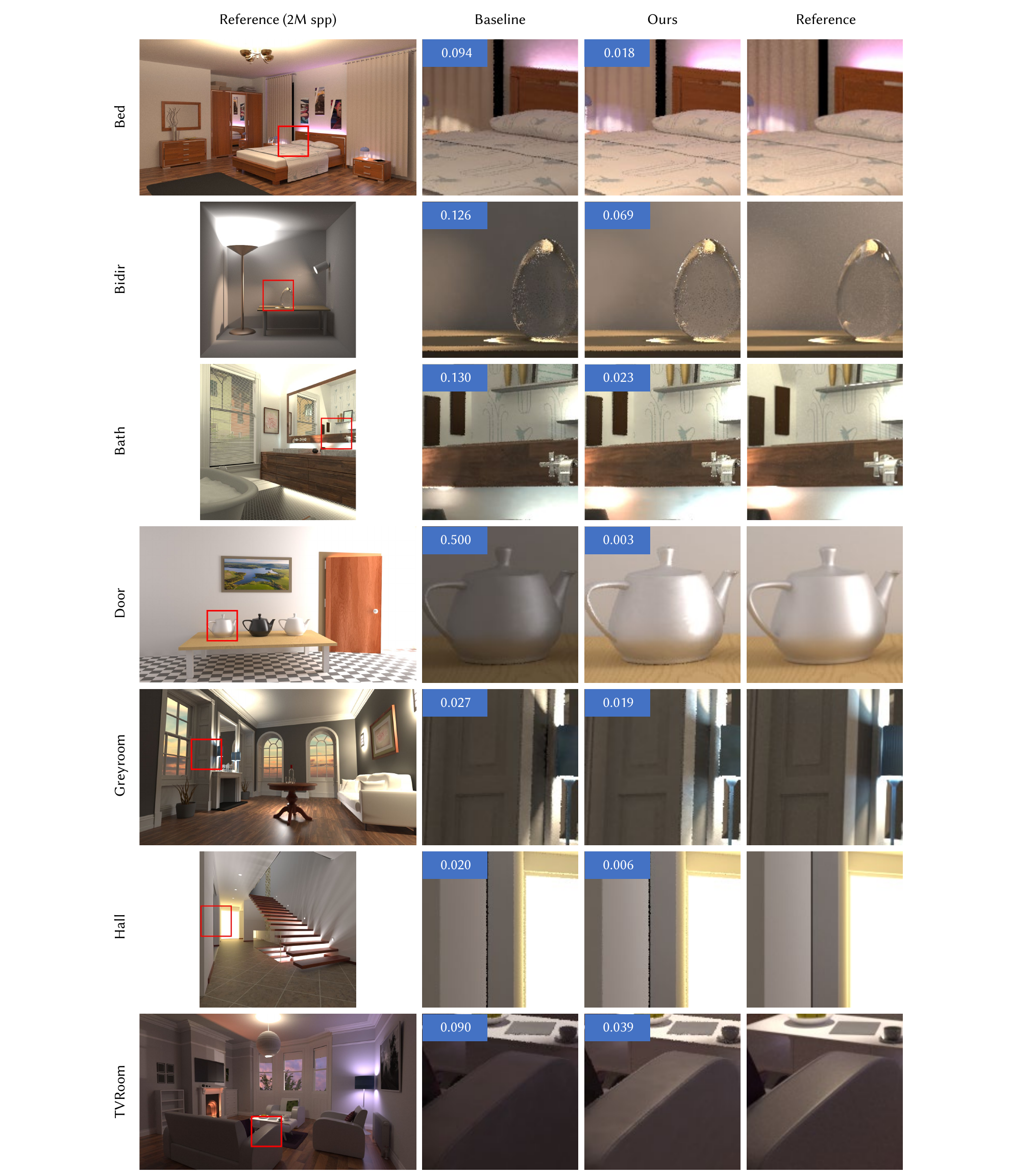}
\caption{
Performance comparison after the same 36K training iterations. The RelMSE error displayed for each image indicates that our method results in a markedly lower whole-image distance toward the reference image. We recommend observing the visual improvements in details related to strong indirect light sources, caustics, reflections, and color bleeding.
}
\label{fig:superior_reconstruction}
\end{figure*}

Additionally, applying the semi-gradient method to generic losses and extending the theory is an important future direction. Since the L2 loss is still widely used in many optimization-based global illumination (GI) fields~\cite{wang2023amortizing, nicolet2023recursive, fischer2023plateau, balint2023joint, deng2022reconstructing, pidhorskyi2022depth}, the theoretical analysis of the semi-gradient method based on L2 is a important contribution to GI. Nevertheless, na\"ively applying the semi-gradient method to generic losses is insufficient to address the bias in gradient estimation. This is because non-L2 losses do not enable unbiased derivative estimation in generel~\cite{nicolet2023recursive}. Despite the limitation, our preliminary experiments (Fig.~\ref{fig:preliminary_results}) show that the semi-gradient technique achieves 2-15 times lower image error compared to the baseline when using root mean square error (RMSE), Huber loss, and mean average error (MAE). These experimental results suggest that our fundamental intuition to ignore differentiation on the RHS, given its faster convergence compared to the LHS, remains valid and beneficial.

\section{Conclusion}
\label{sec:conclusion}

This paper discusses an optimization technique based on partial derivatives to achieve faster and more accurate convergence in residual-based neural rendering methods. We start our discussion with the observation that the right-hand side (RHS), which is the sum of emitted and reflected radiance in the rendering integral, converges significantly faster than the left-hand side (LHS). By deviating from the traditional objective function that pulls the LHS and RHS together, we employ a semi-gradient method that intentionally ignores updates to the RHS for neural cache training. This method achieves an average of 8.8 times lower image error across various scenes compared to existing methods, and reduces per-iteration time by 25-30\% due to avoiding automatic differentiation on the RHS. Through several key proofs, we demonstrate that the semi-gradient loss achieves the same minimum as the existing baseline (i.e., the solution to the rendering equation), while being free from the bias and high variance issues that complicate accurate gradient estimation of the baseline loss. The empirical results show that this characteristic significantly improves training efficiency and rendering accuracy.

\begin{acks}
The authors would like to thank the anonymous reviewers for their valuable comments and insightful suggestions. We also express our gratitude to Kyu Beom Han (KAIST) and our colleagues Jongho Park, Jaeseung Park, Jaeyoung Hwang, Taehong Moon, Byeong-Uk Lee, Minkyu Kim (KRAFTON) for their support and contributions.
We also thank the following people for providing test scenes: \citet{bitterli16rendering}, SlykDrako (Bed), Benedikt Bitterli (Bidir, Door), Mareck (Bath), Wig42 (Greyroom)\footnote{\url{https://www.blendswap.com/blend/13552}}, NewSee2l035 (Hall)\footnote{\url{https://www.blendswap.com/blend/6304}}, and Jay-Artist (TVRoom)\footnote{\url{https://www.blendswap.com/blend/5014}}.
Special thanks to the KRAFTON's KITCHEN~35 team for their invaluable nourishment during this project.
Jaewoong Cho is the corresponding author of this paper.
\end{acks}

\pagebreak 

\bibliographystyle{ACM-Reference-Format}
\bibliography{00_main}

\appendix
\section{Convergence of Gradient Descent for the Semi-Gradient Loss}
\label{appendix:convergence}
We first reformulate the rendering equation (Eq.~(1) of the main report) using the light transport operator \(T\) for a concise derivation:
\begin{equation}
    L^* = E + T \circ L^*,
\end{equation}
where \(T\) takes an outgoing radiance field, proceeds ray tracing, integrals scattering events, and outputs a reflected radiance field.

For a physically correct scene, $T$ is a contraction mapping satisfying:
\begin{equation}
\exists \lambda \in (0, 1): \forall L, \| T \circ L \| \leq \lambda \| L \|.
\end{equation}
That is, roughly speaking, for an arbitrary radiance field \(L\), the energy of the reflected radiance field reduces from that of \(L\) (i.e., energy absorption)~\cite{neumann1995radiosity}.

Then, we can also reformulate the semi-gradient loss using \(T\):
\begin{equation}
    \mathcal{L}_{\sf SG}(L) = \|L - \text{sg}(E + T \circ L)\|^2,
\end{equation}
where we denote the stop-gradient operator as $\text{sg}(\cdot)$ and omit the normalization.

Taking the gradient with respect to $L$, we have:
\begin{align}
    \nabla_L\mathcal{L}_{\sf SG}(L) &= 2(L - E - \text{sg}(T \circ L)) \\
    &= 2(L - E - T \circ L) \quad \text{(in evaluation)}
\end{align}

Consider gradient descent with learning rate $\alpha_{k\geq1} > 0$ for any initial radiance field $L_0$. The update rule for \(k\geq1\) is:
\begin{equation}
    L_k = L_{k-1} - \alpha_k \nabla \mathcal{L}_{\sf SG}(L_{k-1})
\end{equation}

Setting $\alpha_k = \frac{1}{2}$ for all $k$, we can show that the gradient descent converges to $L^*$ for any $L_0$.

The key insight is that $T$ is a contraction mapping. This guarantees that repeated applications of $T$ shrink the error towards zero. Mathematically, we have:
\begin{align}
    \|L_k - L^*\| &= \left\| \left(L_{k-1} - \frac{1}{2} \nabla \mathcal{L}_{\sf SG}(L_{k-1})\right) - (E + T \circ L^*)\right\| \\
    &= \|L_{k-1} - L_{k-1} + E + T \circ L_{k-1} - E - T \circ L^*\| \\
    &= \|T \circ (L_{k-1} - L^*)\| \\
    &= \|T^k \circ (L_0  - L^*)\|.
\end{align}

By the contraction mapping property:
\begin{align}
    \|L_k - L^*\| &= \|T^k \circ (L_0  - L^*)\| \\
    &= \|T\circ T^{k-1} \circ (L_0  - L^*)\| \\
    &\leq \lambda \|T^{k-1} \circ (L_0  - L^*)\|
\end{align}

Continuing in a recursive manner, we get:
\begin{align}
&\leq \lambda^k  \|L_0  - L^* \|
\end{align}

Therefore, as $k$ approaches infinity, the error $\|L_k - L^*\|$ goes to zero, implying convergence to $L^*$:
\begin{equation}
\lim_{k \to \infty}  \|L_k - L^* \| = 0
\end{equation}

Consequently, we find a sequence of gradient descent along $\nabla \mathcal{L}_{\sf SG}$ that converges to $L^*$, which is the solution of the rendering equation, regardless of the starting point $L_0$. $\hfill\blacksquare$

\section{Biasedness of the Baseline Gradient Estimator}
\label{appendix:baseline}
To prove the biasedness of the baseline gradient estimator, we need to show that the expectation of the gradient estimator \(\mathbb{E}[\nabla_\theta \langle \mathcal{L}(\theta)\rangle_{\sf NR}]\) is \textit{not equal} to the true gradient \(\nabla_\theta \mathcal{L}(\theta)\) at some \((x, \omega)\) and \(\theta\). Let us first derive the true gradient as follows:
\begin{equation}
    \begin{split}
    \nabla_\theta \mathcal{L}(\theta) = \nabla_\theta L_\theta \cdot \frac{2(L_\theta - R_\theta )}{\|L_\theta\|^2 + \epsilon} + \nabla_\theta R_\theta \cdot \frac{-2(L_\theta - R_\theta )}{\|L_\theta\|^2 + \epsilon}.
    \end{split}
\end{equation}

Also, by Eq.~(6) of the main report, the expectation of the baseline gradient estimator is:
\begin{equation}
    \begin{split}
    \mathbb{E}[\nabla_\theta \langle \mathcal{L}(\theta)\rangle_{\sf NR}] &= \nabla_\theta L_\theta \cdot \frac{2(L_\theta - \mathbb{E}[\langle R_\theta \rangle])}{\|L_\theta\|^2 + \epsilon} 
    \\
    &\quad + \frac{-2(\mathbb{E}[\nabla_\theta \langle R_\theta \rangle] \cdot L_\theta - \mathbb{E}[\nabla_\theta \langle R_\theta \rangle \cdot \langle R_\theta \rangle])}{\|L_\theta\|^2 + \epsilon}, \\
    &= \nabla_\theta L_\theta \cdot \frac{2(L_\theta -  R_\theta)}{\|L_\theta\|^2 + \epsilon}
    \\
    &\quad + \frac{-2(\nabla_\theta R_\theta \cdot L_\theta - \mathbb{E}[\nabla_\theta \langle R_\theta \rangle \cdot \langle R_\theta \rangle])}{\|L_\theta\|^2 + \epsilon}.
    \end{split}
    \label{eq:biasedness_proof}
\end{equation}
Considering the difference between the above two equations, it suffices to show that the inequality \( \nabla_\theta R_\theta \cdot R_\theta \neq \mathbb{E}[\nabla_\theta \langle R_\theta \rangle \cdot \langle R_\theta \rangle]\) for proving the biasedness (i.e., \(\nabla_\theta \mathcal{L}(\theta) \neq \mathbb{E}[\nabla_\theta \langle \mathcal{L}(\theta)\rangle_{\sf NR}]\)).

\vspace{1em}
Let us start by proving that \( \nabla_\theta R_\theta \cdot R_\theta = \mathbb{E}[\nabla_\theta \langle R_\theta \rangle \cdot \langle R_\theta \rangle]\) if and only if the variance of the RHS estimate is constant (for every \(\theta\)). Then, we will claim that the variance of the RHS estimate cannot be constant, which negates the equality, proving the biasedness of the gradient estimator. 

Suppose that the variance of the RHS estimate is constant for \(\theta\). That is, there exists a constant \(c \in \mathbb{R}\), such that \(\text{Var}[\langle R_\theta \rangle]=c\) at every \(\theta\). Therefore,
\begin{equation}
    \begin{split}
    \nabla_\theta \left (\text{Var}[\langle R_\theta \rangle] \right ) &= 0, \\
    \nabla_\theta \left (\mathbb{E}[\langle R_\theta \rangle ^2] - \mathbb{E}[\langle R_\theta  \rangle]^2\right ) &= 0.
    \end{split}
\end{equation}
By the Leibniz integral rule, we can interchange the differentials and expectations (i.e., integrals) in the last equation:
\begin{equation}
    \begin{split}
    \mathbb{E}[2 \nabla_\theta \langle R_\theta \rangle \cdot \langle R_\theta \rangle] - 2 \nabla_\theta \left( \mathbb{E}[\langle R_\theta \rangle] \right) \cdot \mathbb{E}[\langle R_\theta \rangle] &= 0, \\
    \mathbb{E}[2 \nabla_\theta \langle R_\theta \rangle \cdot \langle R_\theta \rangle] - 2 \nabla_\theta R_\theta \cdot R_\theta &= 0.
    \end{split}
\end{equation}
Thus we proved the one direction. 

The converse is also true. If \(\nabla_\theta \left (\text{Var}[\langle R_\theta \rangle] \right ) = 0\) for every \(\theta\), then there does not exist two parameters, \(\theta_1\) and \(\theta_2\), such that \(\text{Var}[\langle R_{\theta_1} \rangle] \neq \text{Var}[\langle R_{\theta_2} \rangle]\) by the mean value theorem. It proves that the variance is constant for \(\theta\) if \( \nabla_\theta R_\theta \cdot R_\theta = \mathbb{E}[\langle \nabla_\theta R_\theta \rangle \cdot \langle R_\theta \rangle]\).

\vspace{1em}
The variance of the RHS estimate indeed is not constant for \(\theta\). Since the variance can be zero when our radiance prediction network produces zeros at all inputs (e.g., \(\theta=0\)), and the variance can evidently be positive with some \(\theta\)s (e.g., Fig.~5 of the main report).

Therefore, the inequality \(\nabla_\theta \mathcal{L}(\theta) \neq \mathbb{E}[\nabla_\theta \langle \mathcal{L}(\theta)\rangle_{\sf NR}]\) holds, which means the gradient of the baseline estimator is biased to the true gradient of the baseline loss. $\hfill\blacksquare$

\section{Unbiasedness of the Dual-buffer Gradient Estimator}
\label{appendix:dual_buffer}

To prove the unbiasedness of the dual-buffer gradient estimator, we need to show that the expectation of the gradient estimator \(\mathbb{E}[\nabla_\theta \langle \mathcal{L}(\theta)\rangle_{\sf DB}]\) is \textit{equal} to the true gradient \(\nabla_\theta \mathcal{L}(\theta)\) at every \((x, \omega)\) and \(\theta\).

First, the dual-buffer loss estimator is:
\begin{equation}
    \langle \mathcal{L}(\theta) \rangle_{\sf DB} = \frac{(L_\theta-\langle R_{\theta} \rangle_X)\cdot(L_\theta-\langle R_{\theta} \rangle_Y)}{\|\text{sg}(L_\theta)\|^2+\epsilon},
\label{eq:estimator_of_dual_buffer_loss}
\end{equation}
where \(\langle R_{\theta} \rangle_X\) and \(\langle R_{\theta} \rangle_Y\) are two uncorrelated estimates of \(R_\theta\).

Then, the expectation of the gradient of the dual-buffer estimator is:
\begin{equation}
    \begin{split}
    \mathbb{E}[\nabla_\theta \langle \mathcal{L}(\theta)\rangle_{\sf DB}] &= \mathbb{E}\left [\nabla_\theta \left ( \frac{\|L_\theta\|^2 - L_\theta \cdot (\langle R_\theta \rangle_X + \langle R_\theta \rangle_Y)}{\|\text{sg}(L_\theta)\|^2 + \epsilon} \right ) \right] \\
    &\quad + \mathbb{E}\left [\nabla_\theta \left ( \frac{\langle R_\theta \rangle_X \cdot \langle R_\theta \rangle_Y}{\|\text{sg}(L_\theta)\|^2 + \epsilon} \right ) \right], \\
    &= \nabla_\theta \left ( \frac{\|L_\theta\|^2 - L_\theta \cdot (\mathbb{E}[\langle R_\theta \rangle_X] + \mathbb{E}[\langle R_\theta \rangle_Y])}{\|\text{sg}(L_\theta)\|^2 + \epsilon} \right ) \\
    &\quad + \nabla_\theta \left ( \frac{\mathbb{E}[\langle R_\theta \rangle_X \cdot \langle R_\theta \rangle_Y]}{\|\text{sg}(L_\theta)\|^2 + \epsilon} \right ), \\
    &= \nabla_\theta \left ( \frac{\|L_\theta\|^2 - 2L_\theta \cdot R_\theta}{\|\text{sg}(L_\theta)\|^2 + \epsilon} \right ) \\
    &\quad + \nabla_\theta \left ( \frac{\mathbb{E}[\langle R_\theta \rangle_X] \cdot \mathbb{E}[\langle R_\theta \rangle_Y]}{\|\text{sg}(L_\theta)\|^2 + \epsilon} \right ).
    \end{split}
\end{equation}
By the Leibniz integral rule, we can interchange the differentials and expectations (i.e., integrals) in the above equations.

The last equality holds since two estimates, \(\langle R_\theta \rangle_X\) and \(\langle R_\theta \rangle_Y\), are supposed to be uncorrelated in the dual-buffer method. Therefore, starting from the last equality,
\begin{equation}
    \begin{split}
    \mathbb{E}[\nabla_\theta \langle \mathcal{L_\text{DB}}(\theta)\rangle] &= \nabla_\theta \left ( \frac{\|L_\theta\|^2 - 2L_\theta \cdot R_\theta}{\|\text{sg}(L_\theta)\|^2 + \epsilon} \right ) + \nabla_\theta \left ( \frac{R_\theta \cdot R_\theta}{\|\text{sg}(L_\theta)\|^2 + \epsilon} \right ), \\
    &= \frac{\nabla_\theta L_\theta \cdot (2 L_\theta - 2 R_\theta) + \nabla_\theta R_\theta \cdot (-2L_\theta)}{\|L_\theta\|^2 + \epsilon} \\
    &\quad + \frac{2 \nabla_\theta R_\theta \cdot R_\theta}{\|L_\theta\|^2 + \epsilon}, \\
    &= \nabla_\theta \mathcal{L}(\theta),
    \end{split}
\end{equation}
which shows that the gradient of the dual-buffer estimator is unbiased to the true gradient of the baseline loss. $\hfill\blacksquare$

\section{Weighted Dual-buffer Method for Ablation}
\label{appendix:weighted}

To investigate how the partial derivative with respect to the RHS of Eq.~(\ref{eq:estimator_of_dual_buffer_loss}) affects the reconstruction quality, we define the following weighted dual-buffer estimator given a non-negative scalar weight \(w\):
\begin{equation}
    \begin{split}
    \langle \mathcal{L}(\theta; w) \rangle_{\sf WDB} &= \frac{\left\|L_\theta - \text{sg}\left(\frac{\langle R_{\theta} \rangle_X + \langle R_{\theta} \rangle_Y}{2}\right)\right\|^2}{\left\|\text{sg}(L_\theta)\right\|^2 + \epsilon} \\
    &\quad + w \cdot \frac{\left(\text{sg}(L_\theta) - \langle R_{\theta} \rangle_X \right)\cdot\left(\text{sg}(L_\theta) - \langle R_{\theta} \rangle_Y \right)}{\left\|\text{sg}(L_\theta)\right\|^2 + \epsilon},
    \end{split}
    \label{eq:wds_loss}
\end{equation}
where \(\langle R_{\theta} \rangle_X\) and \(\langle R_{\theta} \rangle_Y\) are two uncorrelated estimates of \(R_\theta\). The gradient of the weighted dual-buffer estimator is derived as follows:
\begin{equation}
    \begin{split}
    \nabla_\theta \langle \mathcal{L}(\theta; w) \rangle_{\sf WDB} &= \nabla_\theta L_\theta \cdot \frac{2 L_\theta - (\langle R_{\theta} \rangle_X + \langle R_{\theta} \rangle_Y)}{\left\| L_\theta \right\|^2 + \epsilon} \\
    &\quad + w \cdot \nabla_\theta \langle R_{\theta} \rangle_X \cdot \frac{-(L_\theta - \langle R_{\theta} \rangle_Y)}{\left\| L_\theta \right\|^2 + \epsilon}\\
    &\quad + w \cdot \nabla_\theta \langle R_{\theta} \rangle_Y \cdot \frac{-(L_\theta - \langle R_{\theta} \rangle_X)}{\left\| L_\theta \right\|^2 + \epsilon}.
    \end{split}
\end{equation}
Therefore, it can modulate the influence of the RHS derivatives on network parameter updates by varying the weight \(w\). We thus refer to it as a \textit{weighted} dual-buffer estimator. At \(w=1\), the gradient of the weighted dual-buffer estimator is equivalent to the gradient of the vanilla dual-buffer estimator, and setting \(w=0\) induces the gradient of the semi-gradient estimator.

\begin{figure}[t!]
\centering
\includegraphics[width=\linewidth]{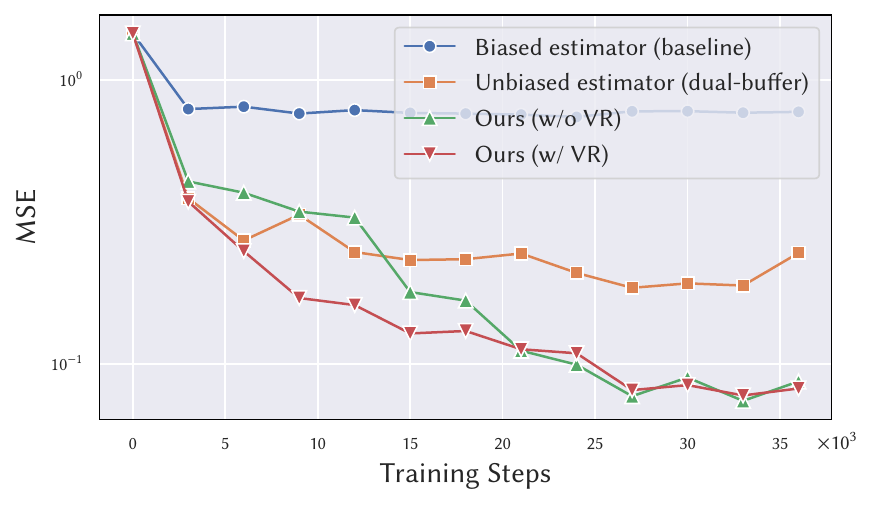}
\caption{
Ablation study on our models with and without an additional path guiding module for variance reduction on the RHS and gradient estimator. We also visualize the dual-buffer method for the reference.
}
\label{fig:converges_faster}
\end{figure}

\end{document}